# Advanced Crash Causation Analysis for Freeway Safety: A Large Language Model Approach to Identifying Key Contributing Factors

**Ahmed S. Abdelrahman**
Corresponding Author
Ph.D. Candidate
Smart and Safe Transportation Lab
University of Central Florida
Email: ahmed.abdelrahman@ucf.edu

**Mohamed Abdel-Aty**
Pegasus Professor and Trustee Chair
Smart and Safe Transportation Lab
Department of Civil, Environmental and Construction Engineering
University of Central Florida
Email: m.aty@ucf.edu

**Samgyu Yang**
Ph.D. Candidate
Smart and Safe Transportation Lab
University of Central Florida
Email: Samgyu.Yang@ucf.edu

**Abdulrahman Faden**
Ph.D. Candidate
UCF Smart & Safe Transportation Lab
Department of Civil, Environment and Construction Engineering
University of Central Florida
Orlando, FL, United States, 32816
Email: Abdulrahman.Faden@ucf.edu
Lecturer
Department of Traffic and Transportation Engineering
Imam Abdulrahman bin Faisal University
Dammam, Saudi Arabia
Email: Akfaden@iau.edu.sa

## ABSTRACT

Understanding the factors contributing to traffic crashes and developing strategies to mitigate their severity is essential. Traditional statistical methods and machine learning models often struggle to capture the complex interactions between various factors and the unique characteristics of each crash. This research leverages large language model (LLM) to analyze freeway crash data and provide crash causation analysis accordingly. By compiling 226 traffic safety studies related to freeway crashes, a training dataset encompassing environmental, driver, traffic, and geometric design factors was created. The Llama3 8B model was fine-tuned using QLoRA to enhance its understanding of freeway crashes and their contributing factors, as covered in these studies. The fine-tuned Llama3 8B model was then used to identify crash causation without pre-labeled data through zero-shot classification, providing comprehensive explanations to ensure that the identified causes were reasonable and aligned with existing research. Results demonstrate that LLMs effectively identify primary crash causes such as alcohol-impaired driving, speeding, aggressive driving, and driver inattention. Incorporating event data, such as road maintenance, offers more profound insights. The model's practical applicability and potential to improve traffic safety measures were validated by a high level of agreement among researchers in the field of traffic safety, as reflected in questionnaire results with 88.89%. This research highlights the complex nature of traffic crashes and how LLMs can be used for comprehensive analysis of crash causation and other contributing factors. Moreover, it provides valuable insights and potential countermeasures to aid planners and policymakers in developing more effective and efficient traffic safety practices.

**Keywords:** Crash causation analysis, Freeways crashes, Large Language Model

## 1. INTRODUCTION

Transportation systems in the United States have been significantly impacted by the country's expanding population and escalating travel demands in recent years. As a result, traffic safety challenges, such as congestion and crashes, have persistently intensified (*1*). In 2023, freeway deaths in the US increased by 9.4% from 2017, reaching nearly 40,990 fatalities NHTSA (*2*). These fatalities and injuries from vehicle crashes place excessive pressure on healthcare systems and the economy, leading researchers, government agencies, and private sector professionals to investigate the causes of crashes and develop strategies to reduce their severity (*3*). A fundamental component of traffic safety studies is the collection of crash data, which traffic authorities typically generate in comprehensive reports using figures, text, and numerical formats after a crash occurs. These reports include interconnected and complex factors such as infrastructure design, human behavior, environmental conditions, drug use, and vehicle issues, making it challenging to evaluate the data and define the precise crash causes (*4*). Researchers have extensively studied the causes of traffic crashes using statistical method (*5-7*). While statistical approaches are easily employed and produce interpretable results, they may not adequately account for the intricate interactions between factors and the distinct features of each crash. These methods face limitations such as assumptions of linearity, underfitting tendencies, and difficulties with missing data (*8*).

To overcome these limitations, machine learning models have been increasingly used for crash analysis. Researchers typically structure the analysis as classification using fixed features extracted from crash reports for summarization and prediction (*9*). Moreover, numerous studies have explored the various factors influencing crash frequency and severity, providing a robust foundation for the development of machine learning models aimed at understanding and predicting crash contributing factors. These factors can be broadly categorized into traffic factors, roadway and geometric factors, driver-related factors, and environmental and weather factors. First, traffic volume has been consistently identified as a significant contributor to increased crash frequency. Studies have shown that higher traffic volumes correlate with higher crash frequencies (*10-12*). Roadway and geometric factors also significantly affect crash frequency and severity. Narrow lanes and sharp curves are commonly associated with higher crash frequencies (*13-15*). For instance, narrower lanes limit the maneuverability of vehicles, thereby increasing the likelihood of crashes (*16*). Especially, driver-related factors, such as demographics and behavior, are critical determinants of crash outcomes. Older drivers are more likely to experience severe crashes, while younger drivers tend to have lower injury risk (*17-19*). Speeding and alcohol use are additional driver behaviors that markedly increase crash injury severity (*20,21*). Other factors, including lane changes, phone use, and aggressive driving, also impact crash frequency and severity (*22,23*). The use of mobile phones while driving significantly contributes to distraction-related crashes (*24*). Lastly, for environmental factors, adverse weather conditions, such as rain and snow, have been shown to increase crash frequencies but sometimes decrease the severity of injuries due to lower driving speeds (*25-27*). Light conditions also play a significant role, for instance, crashes occurring in daylight tend to be less severe compared to those occurring in darkness (*28,29*).

Although, machine learning models provide more accurate and valuable outcomes compared to traditional methods because they handle both linear and nonlinear assumptions and address the overfitting challenge, they are more difficult to comprehend and may oversimplify the data, failing to provide in-depth explanations of specific crash occurrence and severity factors (*30*). Despite advancements in these methods, there remains a need for user-friendly tools that make understanding crash factors easy and efficient for non-experts, such as planners, policymakers, and other entities. Recently, the large language models (LLMs), which are powerful artificial intelligence approach that interpret and generate natural language, making them highly interactive, have proposed and utilized in various domain such as medicine (*31*), Social Sciences (*32*), law (*33*), Computational Biology (*34*). Their "large" designation is justified by their extensive training on massive amounts of text data and the integration of billions or trillions of parameters from different domains and topics in the internet, enabling them to perform complex language tasks (e.g., question-answering, summarization, text completion, emotion analysis, and problem-solving) with exceptional accuracy. The OpenAI (ChatGPT) is one of the transformer-based pre-trained LLMs that provide powerful wide content generation that closely reflects human abilities, readable outcomes, and universal knowledge language (*35*). Moreover, LLMs can be fine-tuned for certain tasks or datasets by training them on smaller, more targeted datasets from specific domains or topics. This specific-domain pre-trained LLMs will optimize their performance in fields such as transportation, finance, medicine, law, etc. Hence, the fine-tuned procedures for specialized tasks or applications underscore the flexibility and the potential of foundation LLM (i.e., Llama models by Meta) (*36*). Consequently, LLMs are more user-friendly interface and has more precise reasoning. With the help of large amounts of training data, it may improve its knowledge base and take on increasingly difficult linguistic tasks (*37*).

By leveraging the extensive research on crash contributing factors, LLMs can capture the complex relationships between these factors and crash outcomes. Such models hold promise for improving traffic safety by providing data-driven insights into crash causation and enabling targeted interventions. Hence, this study proposes using LLMs for analyzing freeway crashes and applying crash causation analysis to better understand the root causes of crashes and how to reduce them accordingly. This approach could address the limitations of current statistical and machine learning methods, enhance the prediction of crash-contributing factors, and improve overall traffic safety practices.

The remainder of the paper is structured as follows: Section 2 highlights the relevant work using LLMs in the traffic domain. Section 3 addresses the database collection. Section 4 illustrates the fine-tuning process of the LLM. Section 5 presents how the fine-tuned LLM will be used for crash causation analysis. Section 6 discusses the findings in depth. The conclusions and suggestions for future work are presented in Section 7.

## 2. RELATED WORK

LLMs have recently been applied in various fields such as medicine and computational biology, efforts to utilize them for analyzing crash data and identifying causes in the field of traffic safety are also emerging. Complex analyses integrate diverse data types such as images or videos. Wu, Li and Xiao (*38*) proposed 'AccidentGPT', which is a foundation model that utilizes multi-modal inputs, such as pre- and post-crash site photos, GPS location, to automatically reconstruct crash and provide comprehensive multi-task analysis. Leveraging LLMs and large multi-modal models, it aims to overcome the limitations of traditional methods which are often constrained by manual processes, subjective biases, and privacy concerns. Significantly, research is utilizing unstructured data, such as explanation and descriptions of crash, to analyze crash and consider contributing factors. Shao, Shi, Zhang, Shiwakoti, Xu and Ye (*4*) demonstrates the effective application of natural language processing (NLP) and machine learning in crash analysis, offering new insights into behavior-cause relationships and injury severity prediction using both of unstructured (crash description) and structed (crash-related variables) data. The integration of LLM techniques significantly enhances the ability to process and analyze complex crash data with 0.79 of accuracy, providing valuable tools for developing targeted road safety interventions. Grigorev, Saleh, Ou and Mihaita *(39)* explores the integration of LLMs, based on BERT (Bidirectional Encoder Representations from Transformers), into machine learning workflows for crash severity classification, demonstrating that LLM-extracted features combined with traditional features significantly improve prediction accuracy, with the F1-score increasing from 0.82 to 0.89 compared to using traditional features alone. This hybrid approach leverages the LLMs' ability to process unstructured textual data, simplifying feature engineering and enhancing the performance of machine learning models like RandomForest and XGBoost across datasets from multiple regions. Fan, Wang, Zhao, Zhao, Ivanovic, Wang, Pavone and Yang *(30)* introduces Crash-LLM, based on Llama-2, utilizing LLMs to predict crash outcomes based on detailed textual crash reports. The LLM-based approach significantly outperforms traditional models, with an average F1-score improvement from 34.9% to 53.8%, and it enables insightful what-if analyses for traffic safety improvement.

The studies reviewed highlight the various uses of language models in traffic safety research, such as reconstructing crashes with multi-modal inputs, predicting injury severity, and classifying crash severity. The results show that incorporating advanced language models greatly improves the accuracy and efficiency of these tasks, underscoring their potential to address numerous challenges within the traffic safety field. Many studies have conducted crash analysis on freeways, but they often lack identification of the direct causes of these crashes. Therefore, this research takes a deeper approach to identify the root causes of crashes on freeways, focusing on specific freeway segments by applying crash causation analysis using LLM.

## 3. DATABASE COLLECTION

Traffic safety studies focusing on freeways have been collected from various top-ranked journals to fine-tune the LLMs. To automate the extraction of the key findings from these collected research studies, ChatGPT-4, for its superior performance in reasoning and in-context learning compared to other LLMs *(40)*, has been utilized to retrieve this information from these published research papers and reformat the outcomes into 'prompt' and 'completion' in JSON format within an array for fine-tuning the LLMs specifically for the traffic safety domain. In this study, 226 traffic safety studies were compiled as a training textual dataset. As shown in **Figure 1**, 26% (59 papers) were from the Accident Analysis & Prevention Journal (AAP), 11% (25 papers) from Analytic Methods in Accident Research (AMAR), 11% (24 papers) from Transportation Research Record (TRR), 7% (15 papers) from the Journal of Safety Research, 4% (10 papers) from Transportation Research Part C, 4% (10 papers) from the Journal of Transportation Safety & Security, and 37% (83 papers) from other top-ranked journals such as IEEE Transactions on Intelligent Transportation Systems, and Transportation Research Part F: Traffic Psychology and Behavior. These studies provided insights into the relationship between crash occurrence and contributing factors related to the environment, drivers, traffic, or geometric design. This fine-tuning process aims to optimize the LLMs for the traffic safety domain, resulting in accurate and reliable knowledge for traffic causality analysis.

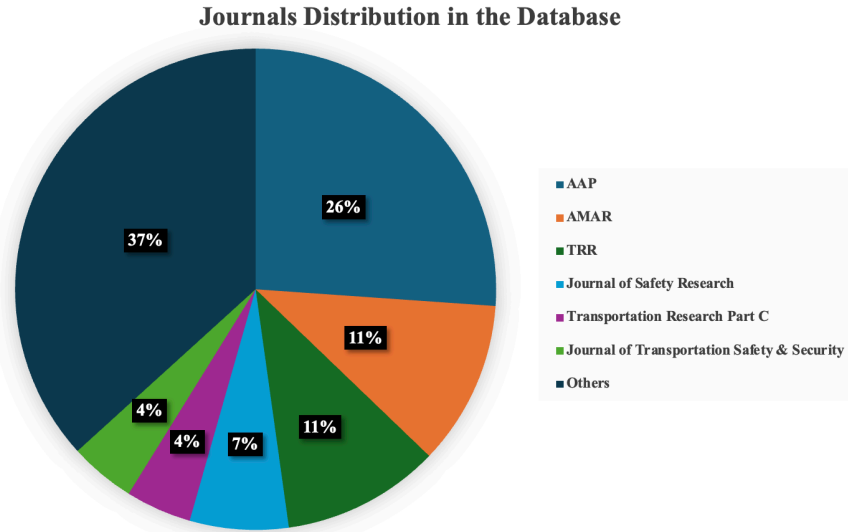

**Figure 1 Distribution of Traffic Safety Studies in Database**

## 4. FINE-TUNING LLM FOR FREEWAY CRASH ANALYSIS

After preparing the database collected from 226 research studies related to crashes on freeways, the database is used to fine-tune an LLM to be more domain-specific regarding freeway crashes and their contributing factors. Subsequently, this LLM is employed to identify the primary causes behind freeway crashes. This task requires high reasoning capabilities to understand all the correlations and causative factors of freeway crashes and accurately identify the main causation for each crash. To achieve this, a comparison was conducted among several LLMs to select the most suitable LLM for this task. **Figure 2** illustrates the comparison among seven LLMs based on the Massive Multi-task Language Understanding (MMLU) score (*41*), which is a benchmark designed to measure knowledge acquired during pretraining by evaluating models exclusively in zero-shot and few-shot settings, and the Artificial General Intelligence Evaluation (AGIEval) dataset (*42*), that assesses LLMs' capabilities in reasoning, understanding, and critical thinking. Based on the scores of these seven LLMs (*43-45*), **Figure 2** demonstrates that the Llama3 model with 8 billion parameters outperforms others in terms of understanding and generating highly reasonable outcomes.

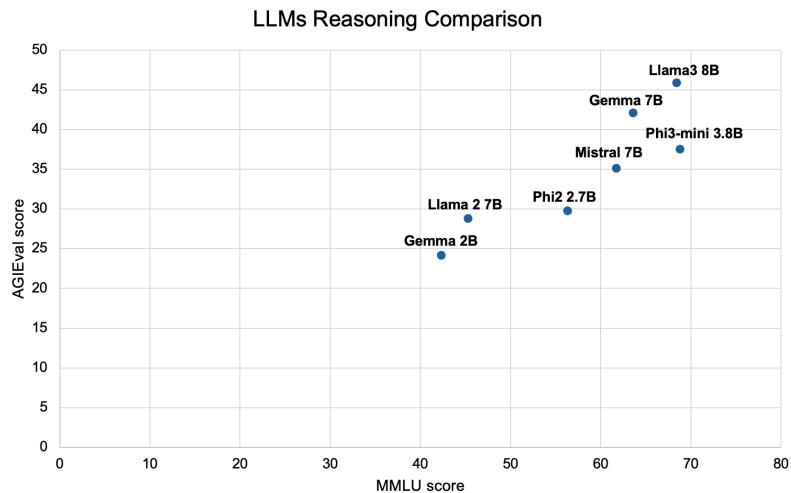

**Figure 2 Comparison between LLMs on reasoning quality**

Given the substantial size of the Llama3 8B model, full fine-tuning is impractical due to its high computational and memory demands. Instead, a more efficient approach is required for the fine-tuning process. One of the most effective methods is the use of adapters. Among the various Adapter-Based Performance Efficient Fine-Tuning (PEFT) techniques for LLMs, the low-rank adaptation (LoRA) method stands out. LoRA often performs on par with or better than full fine-tuning and surpasses other adapter-based methods for fine-tuning large models (*46*). Additionally, LoRA helps prevent the model from overfitting to the fine-tuning data and retains the model's original knowledge by minimizing catastrophic forgetting compared to full fine-tuning (*47*). To further reduce the computational requirements for the fine-tuning process, the base model will undergo quantization. Quantization is a technique that reduces the precision of the numbers representing model parameters and activations, converting them from higher bit-width formats (e.g., 32-bit floats) to lower bit-width formats (e.g., 8-bit integers). This significantly decreases the model's memory footprint and computational requirements, enabling faster and more efficient inference. QLoRA (Quantized Low-Rank Adapter) utilizes this approach to facilitate efficient fine-tuning of LLMs on limited hardware. By employing 4-bit NormalFloat (NF4), an optimized data type for normally distributed weights, and double quantization, which further reduces memory usage by quantizing the quantization constants, QLoRA drastically cuts down the memory needed without sacrificing performance (*48*). **Figure 3** illustrates the complete pipeline for the fine-tuning process of Llama3 8B using QLoRA. The first step involves loading the created database in JSON format, which contains question-and-answer pairs based on key findings from over 200 research studies. Next, the base model (Llama3 8B) is loaded, followed by the tokenization of the entire dataset, as LLMs process words in the form of tokens. The QLoRA preparation involves quantizing the base model from FP32 to NF4, setting the LoRA parameters, running the fine-tuning process, and finally merging the adapters with the base model to prepare it for use. During fine-tuning with QLoRA, the base model's parameters remain frozen (unchanged), and only the adapters are updated. **Figure 4** presents the training and evaluation loss during the fine-tuning of the Llama3 8B model using the QLoRA technique, conducted on an Nvidia A100 40 GB GPU and and took approximately 1 hour to finish the fine-tuning. The plot shows the learning curves over 180 training steps, with the y-axis representing loss values and the x-axis denoting the number of steps. Initially, both the training and evaluation losses start at approximately 2.5, indicating high initial error rates. As training progresses, the losses decrease sharply, stabilizing around 1.25 after approximately 60 steps. Beyond this point, the losses continue to decline gradually, with both curves running closely parallel, indicating consistent performance between training and evaluation phases. The minimal divergence between the training and evaluation loss curves suggests that the model is not overfitting and is generalizing well on the evaluation dataset.

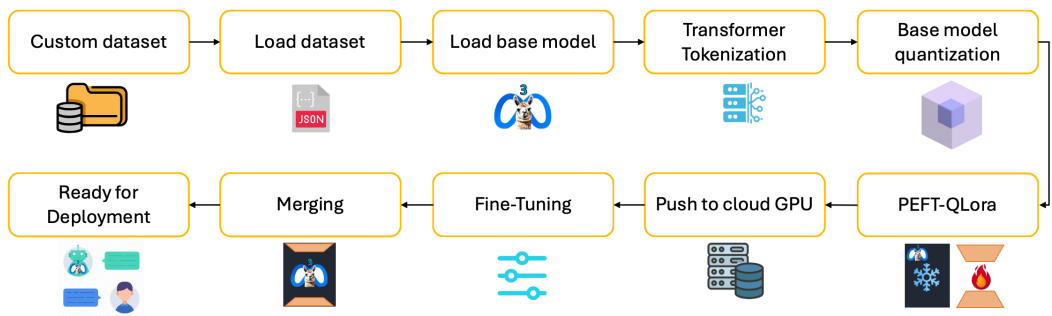

**Figure 3 Pipeline of fine-tuning Llama3 8B using QLoRA**

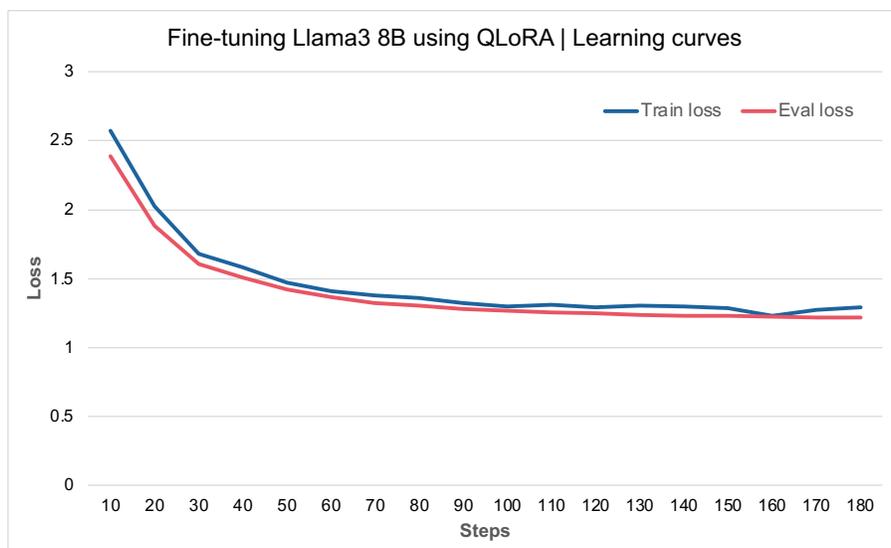

**Figure 4 Training curves of Llama3 8B fine-tuning using QLoRA on domain-specific data (freeways crash studies)**

## 5. CRASH CAUSATION ANALYSIS USING LLM

After fine-tuning the Llama3 8B model using QLoRA on a dataset incorporating 226 research studies, it has become more familiar with crash studies on freeways, the contributing factors, and the influence of these factors. The fine-tuned Llama3 will be employed to identify the main causes of crashes. As presented in **Table 1**, crash parameters are imported from three sources: Signal4Analytics, RITIS Traffic Data, and RITIS Event Data. Signal4Analytics (S4A) is a statewide web-based interactive crash mapping and analysis system. In the crash data, contributing factors that are associated with crash occurrences in previous studies, such as speeding-related crashes, are considered. Traffic and event data are collected through RITIS (Regional Integrated Transportation Information System), a situational awareness, data archiving, and analytics platform used by transportation officials (*49*). The traffic data includes variables such as speed, traffic volume, and occupancy, which are aggregated on a segment basis. This data is used to represent traffic conditions at the time of the crash and just before it occurred. Event data captures unusual situations near the crash site, such as maintenance work and traffic congestion, which are not typically recorded in crash and traffic data. By including event data, we can achieve a detailed consideration of the road environment around the crash location, capturing special conditions that might influence crash occurrences (*50*). While previous traffic safety studies frequently utilized traffic and crash data, they rarely considered unusual road environments like event situations (*51,52*). This inclusion of event data provides a more comprehensive analysis of the factors influencing crashes, addressing gaps in traditional studies and offering deeper insights into the dynamics of crash occurrences.

Although the LLM was trained on information extracted from research studies, it wasn't trained to classify crash causation. Therefore, the fine-tuned LLM will utilize zero-shot classification to determine crash causation. Zero-shot classification allows models to categorize data into unseen classes without needing labeled examples, making it particularly useful when ground truth data is unavailable (*53,54*). LLMs like Llama models by Meta excel in this area due to their extensive pre-training on diverse datasets (*55*). Additionally, fine-tuning on the collected database has provided the LLM with condensed information about freeways, enabling it to act as an expert in identifying possible crash causation given crash parameters.

**Table 1 List of crash parameters**

| Data | Parameters | Description |
|------|-----------|-------------|
| **S4A Crash Report** | Crash Date and Time | Crash data and time |
| | Crash Type | Type of crash (i.e, rear-end, sideswipe) |
| | Crash Severity | 5-level of crash severity (Fatal, Incapacitating Injury, Non-incapacitating injury, Possible injury, No injury) |
| | Light Condition | Light (Daylight, Dark with light, Dark without light, Others) |
| | Weather Condition | Weather (Clear, Cloudy, Rain, Others) |
| | Road Surface Condition | Surface condition (Dry, Wet, Others) |
| | Roadway Alignment | Alignment (Straight, Curve, Others) |
| | Is Speeding Related | Crash is related with speeding or not (Yes, No) |
| | Is Aggressive Driving | Crash is related with aggressive driving or not (Yes, No) |
| | Driver Age | Driver's age at time of crash |
| | Driver Condition | Driver's condition at time of crash (normal, under the influence of alcohol/drug, others) |
| **RITIS Traffic Data** | Speed of Roadway Segment at a Time of Crash | Speed of roadway segment at a time of crash (High: Over 90% of Speed Limit, Medium: 60-90% of Speed Limit, Low: Less than 60% of Speed Limit) |
| | Level of Service of Roadway Segment at a Time of Crash | Level of service at time of crash (LOS A - F) |
| | Speed of Roadway Segment 5 Minutes Before Crash | Speed of roadway segment 5 minutes before the crash (High: Over 90% of Speed Limit, Medium: 60-90% of Speed Limit, Low: Less than 60% of Speed Limit) |
| | Level of Service of Roadway Segment 5 Minutes Before Crash | Level of service 5 minutes before the crash (LOS A - F) |
| **RITIS Event Data** | Event Near Crash Location | Event within 1-mile from crash location at time of crash (No event, Road maintenance operations, Traffic congestion, others) |

**Figure 5** presents the pipeline of deploying the fine-tuned Llama3 8B for crash causation. The crash factors imported from Signal4Analytics and RITIS are combined into one full-text representation, which is then be tokenized, as LLMs operate on words as tokens. Tokenization is the process of breaking down a text sequence into smaller tokens, called tokens, which can be words, sub words, or characters. In the context of Natural Language Processing (NLP), tokenization is crucial as it transforms raw text into a format that machine learning models, especially LLMs, can process (*56-58*). It enables these models to understand and generate text by converting input data into manageable and standardized units. After tokenizing the full-text representation of the crash parameters, the fine-tuned Llama3 8B will be asked to identify the crash causation. Additionally, due to its highly generative capabilities, the LLM will provide an explanation for this causation to support the crash causation classification. This explanation is essential to identify other significant contributing factors of the crash, highlighting the root causes of crashes at each location. It also suggests effective countermeasures to improve safety and reduce the number of crashes at each location.

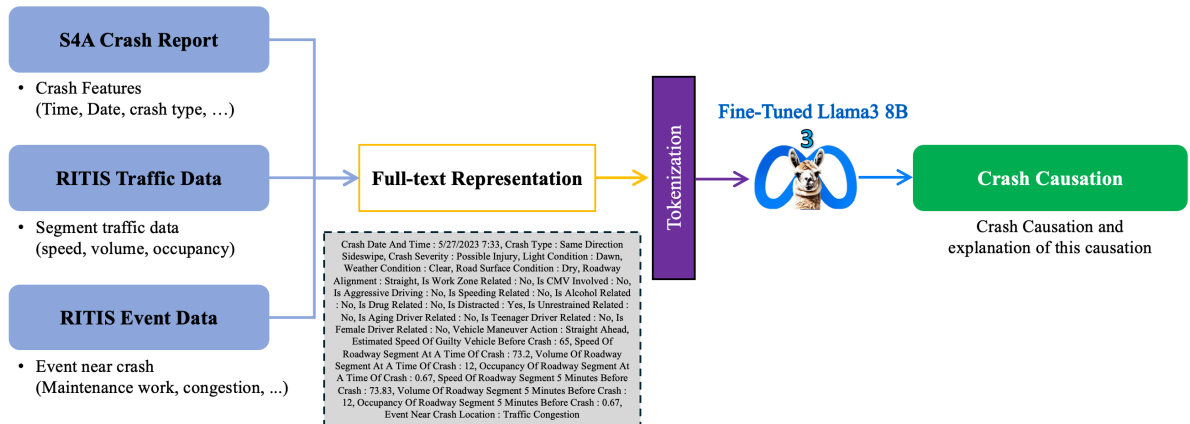

**Figure 5 Diagram of utilizing the fine-tuned Llama3 8B in crash causation analysis**

## 6. RESULTS AND DISCUSSION

### 6.1 Crash Causation Analysis

Using Llama3 8B after training on more than 200 studies on traffic safety and influencing factors, key crash causes for specific cases, which are actual crashes obtained from S4A, were inferred. The LLM's results provide both the main causes and additional explanations for the crash situations **(Figure 6)**.

In Example #1, at the time of the crash, the driver was under the influence of alcohol, which was identified as a major cause of the crash. The explanation of the crash includes several factors. First, the driver's abnormal state was a primary cause of the crash. Driving under the influence of alcohol delays reaction times and impairs judgment, significantly increasing the likelihood of a crash (*21,22,29*). Second, the crash occurred during road maintenance operation, which further increased the risk of the crash. Road maintenance can narrow or create irregularities in the roadway, making driving conditions more challenging (*59,60*). Third, the dark lighting conditions and wet road surface limited the driver's visibility and increased stopping distances, thereby heightening the risk of a crash (*28,29*). Fourth, the high speed of the roadway segment at the time of the crash also contributed to the severity of the crash. High speeds increase the energy during a collision, leading to more severe damage (*20,22,61*). Therefore, this example provides a reasonable and consistent analysis of the causes and preventive measures based on the given information. It concludes that effective counter measures to reduce alcohol-impaired driving and improving road maintenance practices can prevent such crashes effectively.

In Example #2, the cause of the crash was determined to be the driver's speeding and aggressive driving. At the time of the crash, the driver was speeding and engaging in aggressive driving behavior. The explanation highlights that the driver's speeding and aggressive driving were the primary causes of the rear-end collision (*61*). The driver was engaging in aggressive driving behaviors, such as sudden braking or changing lanes without signaling, which created dangerous conditions for other drivers and led to the crash. Additionally, the crash occurred on a straight road segment during daylight hours with clear weather conditions. The road surface was dry, and there were no significant events or obstacles near the crash location. Furthermore, the driver was 27 years old and in a normal condition at the time of the crash. Speeding and aggressive driving are well-known primary causes of crashes, supported by numerous studies (*20,22,29*). The fact that the crash occurred in clear weather conditions during daylight hours further emphasizes that the driver's behavior was the main cause of the crash. As such, this example also provides a reasonable and consistent analysis of the causes and preventive measures of the crash.

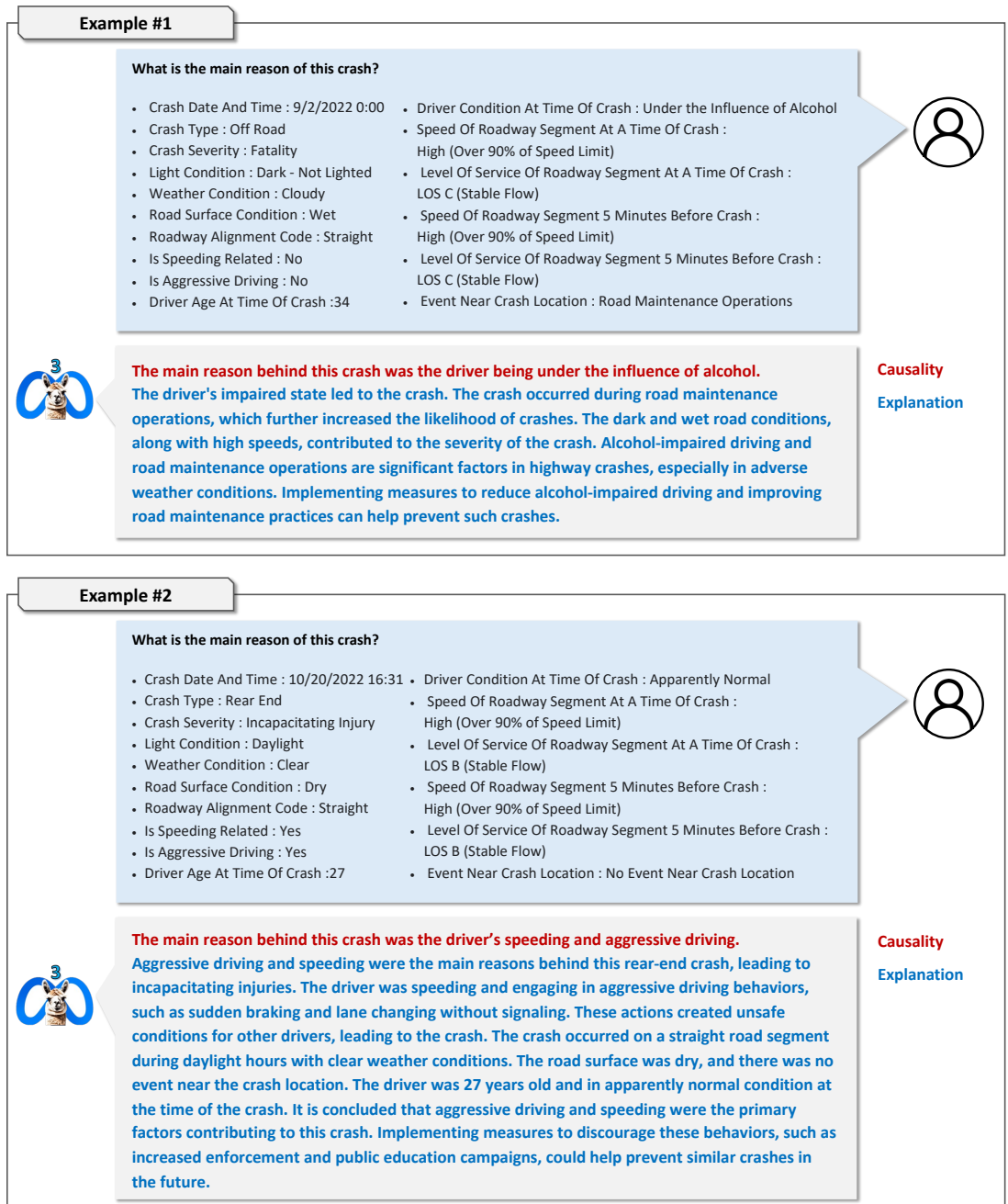

**Figure 6 Examples of crash causation along with explanation**

To evaluate the appropriateness of these results, a questionnaire was conducted with six researchers whom their study is directly related to the field of traffic safety. The questionnaire consisted of three questions:

1. **Is Causation Reasonable?**: This question aims to evaluate whether the 'causation' is reasonable and matches with the literature. Specifically, it seeks to determine if the presented causes are plausible and credible as actual crash causes.
2. **Is Explanation Reasonable?**: This question focuses on whether the 'explanations' are logical, consistent, and reliable. It examines how well the explanations are connected to the causes and whether the content is appropriate for crash analysis.
3. **Is Explanation Descriptive and Clear?**: This question assesses whether the 'explanations' are written clearly and understandably, and if they include sufficient detail. It evaluates if the explanations are natural, easy to understand, and provide the necessary information in a comprehensible manner.

Through these questions, the questionnaire aims to determine whether the results are practically valid and understandable according to the perspectives of traffic safety professionals. Each researcher evaluated the generated crash causation analysis on six different crashes similar to and including the ones shown in **Figure 6**. The results of the questionnaire were that 88.89% agreed with the crash causation, highlighting the main crash influencing factor, and 86.11% voted that the explanation is reasonable and supports the crash causation clearly. This high level of agreement from traffic safety researchers supports that the LLM's results are reliable and insightful. The ability to identify and explain crash causation effectively can greatly enhance our understanding and prevention of crashes. Moreover, it underscores the potential for such models to be valuable tools in crash analysis and prevention, offering a comprehensive view that includes both direct and contextual factors influencing crashes.

**6.2 Crash Hot-spot Segments Analysis**

In 2023, using the data collected through S4A, 1,631 crashes were identified on I-4 in Florida's District 5. Based on this data, segments were designated as dangerous segments where crashes frequently occurred or where severe crashes were common **(Figure 7)**. This chapter categorizes and analyzes the causes of crashes in each of these dangerous segments, utilizing the fine-tuned Llama3 8B on the created database through zero-shot classification of the main causation of each crash on each segment.

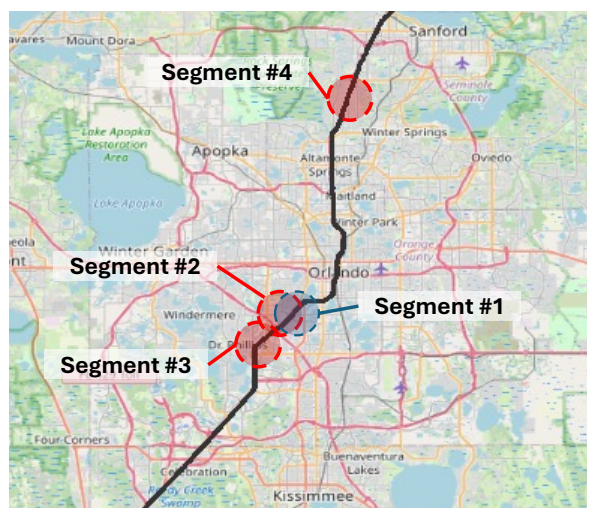

**Figure 7 Crash Hot-spot Segments in I-4**

*6.2.1 Analysis of Segment 1*

Segment 1 had 62 crashes in 2023, of which the analysis was performed based on 46 crashes that had accompanying traffic data. As shown in **Figure 8,** The crash causation analysis in Segment 1 revealed that traffic congestion was found to be the most frequent related factor, indicating that it is a significant underlying issue (*10,11*). Driver inattention is intricately linked to traffic congestion. It is essential to recognize that traffic congestion is not only a direct problem of crashes but also exacerbates other factors. Driver inattention, responsible for 78% of crashes, often arises from the monotonous nature of congested driving conditions. The slow pace and frequent stops can lead to disengagement or distraction, such as using mobile devices, conversing, or other in-car activities. This lack of attention is especially hazardous in congested traffic, where sudden stops or slowdowns are common. Risky driving behaviors, noted in 13% of crashes, are less significant but still important. These behaviors, possibly related to the age of the driver (e.g., teenagers and females), include unsafe maneuvers like lane changing or speeding to bypass slower traffic. Such actions not only increase crash risk but also disrupt traffic flow. Given the high proportion of crashes linked to traffic congestion, addressing this issue is vital for enhancing safety in Segment 1. Prioritizing congestion management strategies, such as implementing ramp metering to regulate flow and improving road infrastructure, is essential.

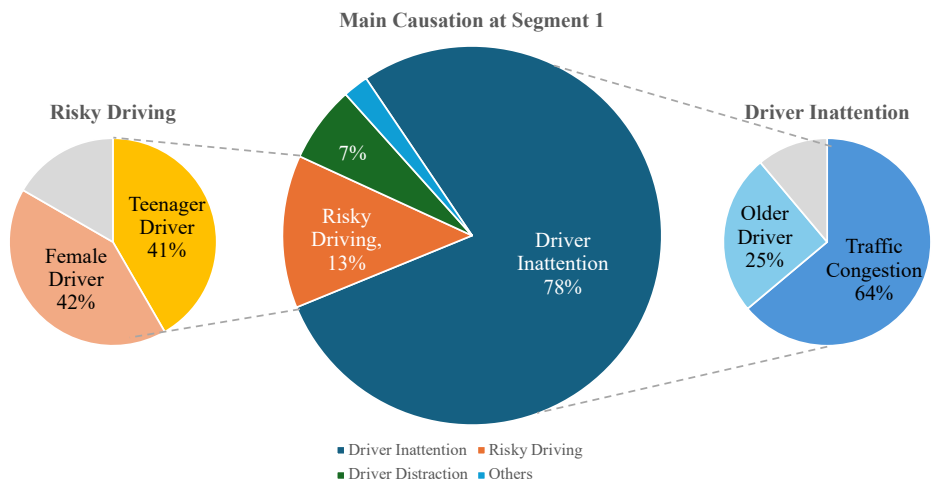

**Figure 8 Categorized Main Causation at Segment 1**

*6.2.2 Analysis of Segment 2*

Segment 2 had 34 crashes, with analysis performed on 28 crashes that had accompanying traffic data (**Figure 9**). Speeding was the most frequent cause, accounting for 43% of crashes, indicating common speeding on highways (*20,29*). This high percentage indicates a prevalent issue of speeding on this segment, which not only directly contributes to crashes but also relates to other factors. For example, driver inattention, the second major cause of crashes at 36%, is often a consequence of or exacerbated by speeding. When vehicles travel at high speeds, drivers have less time to react to changing road conditions, making inattentive driving more dangerous, particularly during maneuvers like lane changes, which account for 60% of these inattention-related crashes. Furthermore, risky driving, noted for 14% of crashes, is closely tied to speeding. Teenage drivers, who were involved in 75% of these risky driving crashes, often exhibit a combination of inexperience and a tendency to speed, heightening the likelihood of severe outcomes (*62*). The interconnectedness of these factors highlights the need for a comprehensive consideration to addressing speeding and its related risks. This could include stricter enforcement, such as installing speed cameras and increasing police patrols to uphold speed limits and implementing dynamic speed limits that adjust based on traffic and weather conditions. Therefore, addressing the issue of speeding in Segment 2 requires a multifaceted strategy that considers its role as both a direct cause and a relating factor for other unsafe driving behaviors.

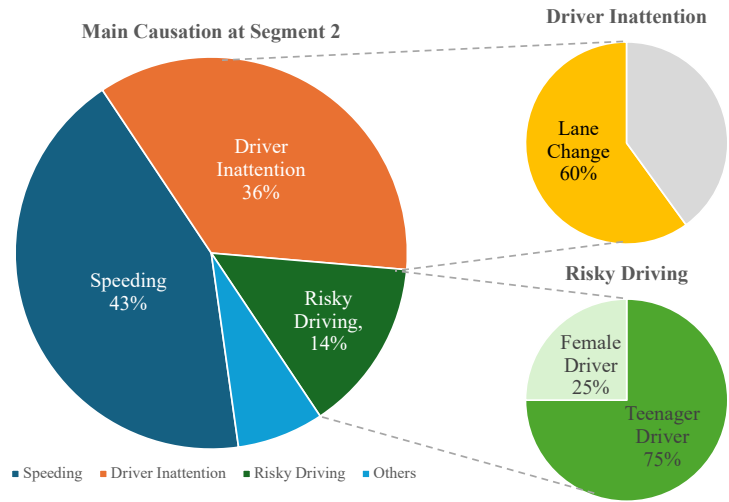

**Figure 9 Categorized Main Causation at Segment 2**

### 6.2.3 Analysis of Segment 3

Segment 3 had 51 crashes, with analysis performed on 27 crashes that had accompanying traffic data (**Figure 10**). Driver distraction was the most frequent cause, accounting for 62% of crashes, often due to cell phone use, navigation system manipulation, and in-vehicle conversations, which reduce driving safety, especially when quick reflexes and constant attention are required. Additionally, driver inattention was the second major cause, accounting for 19% of crashes, with 40% occurring during lane changes. Risky driving accounted for 11% of crashes, often due to speeding or aggressive driving (*20,22*). To address these issues comprehensively and enhance road safety in Segment 3, Advanced Driver Assistance Systems (ADAS) can play a crucial role in reducing crashes caused by driver distraction and inattention. Systems such as lane departure warnings, adaptive cruise control, and automatic emergency braking provide real-time alerts and corrective actions, helping drivers stay focused and avoid crashes. Integration with smartphone systems can enable hands-free functionality, minimizing distractions from mobile devices. Furthermore, Intelligent Transportation Systems (ITS) and Variable Message Signs (VMS) are beneficial in this context. Deploying VMS can offer real-time information about traffic conditions, speed limits, and potential hazards, thereby addressing both distraction and inattention. These signs can alert drivers to slow down, focus, or prepare for upcoming road conditions, such as roadworks or adverse weather, allowing them to adjust their driving accordingly. By integrating these technologies and strategies, a comprehensive approach can be developed to decrease the risks related to driver distraction and inattention.

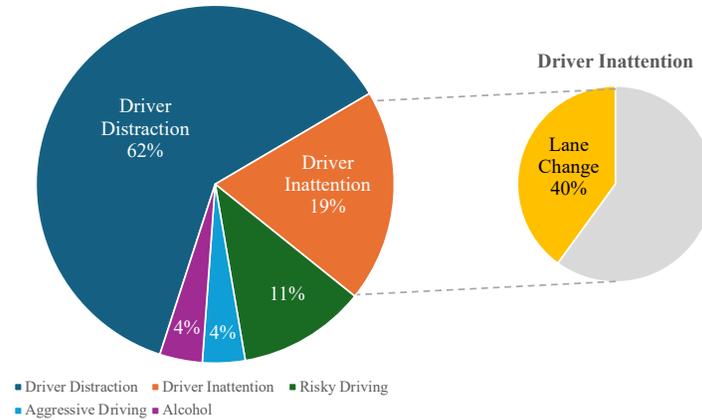

**Figure 10 Categorized Main Causation at Segment 3**

*6.2.4 Analysis of Segment 4*

Segment 4 had 13 crashes in 2023, with analysis performed on 11 crashes that had accompanying traffic data (**Figure 11**). Crashes involving commercial vehicles occur more frequently compared to other segments and tend to result in severe injuries. These crashes are often linked to driver fatigue or distraction (*14,63*). The segment is located in a suburban area and features a straight road, which can lead to a monotonous driving environment. This monotonous setting can cause drivers to lose focus and become fatigued, increasing the risk of crashes. Additionally, speeding is a significant issue in this area, accounting for 37% of all crashes. Driver fatigue contributes to 27% of crashes, while driver distraction is responsible for 18%. Therefore, to improve safety in Segment 4, enhanced speed limit enforcement and monitoring are beneficial. Also, to combat driver fatigue, it is important to provide adequate rest areas and rest stops along the route. These facilities can offer drivers a place to rest and recharge, especially on long stretches of road. Additionally, separating commercial vehicles from regular traffic through protective barriers or designated lanes can reduce the severity of crashes, minimizing interactions between large commercial vehicles and smaller passenger vehicles.

In summary, the results demonstrate that crash causation using LLM effectively identify the primary causes of crashes in high-risk segments of the I-4 highway. By accurately pinpointing these causes, the research offers valuable insights for proposing effective traffic safety measures. These targeted interventions can significantly contribute to preventing crashes and enhancing overall road safety.

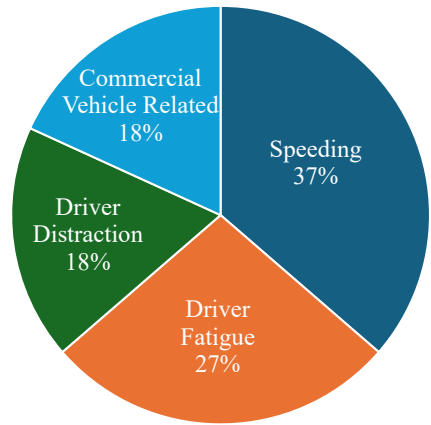

**Figure 11 Categorized Main Causation at Segment 4**

## 7. CONCLUSIONS AND FUTURE WORK

Understanding the main factors behind crashes and developing strategies to mitigate their severity is crucial towards reducing crashes and improving road safety. This study leveraged large language model (LLM) to analyze freeway crash data, identify contributing factors, and provide comprehensive explanation of the main crash causation. The research involved creating a database from 226 traffic safety studies from top-ranked journals related to traffic safety, creating a comprehensive training dataset that includes analyses about the correlations between the environmental, driver, traffic, and geometric design factors and crashes on freeways. Llama3 8B model was fine-tuned using QLoRA on the database to be expert in freeways crashes studies enabling accurate crash causation analysis through zero-shot classification. Unlike traditional methods, the fine-tuned LLMs can handle complex interactions and in-context learning, improving accuracy and providing detailed, interpretable explanations for the crash causation analysis. The conducted analysis on four freeway segments shows that driver inattention, speeding, and distraction are the primary causes of crashes across all segments. Traffic congestion and elderly drivers increase inattention driving condition, while risky behaviors are prevalent among female and teenage drivers. This research demonstrates the effectiveness of LLMs in crash causation analysis and offered valuable insights for planners and policymakers for more effective solutions to reduce crashes on freeways. Driving assistance systems are highly recommended to enhance road safety. Blind spot detection can significantly reduce the risk of crashes during lane changes, while emergency braking is especially effective in preventing collisions during traffic congestion. Future work incorporates expanding the database to include large handbooks as NCHRP and AASHTO, additionally, crash causation analysis can be applied on further other road types and junctions such as intersections.

## AUTHOR CONTRIBUTIONS

The authors confirm contribution to the paper as follows: study conception and design: A. Abdelrahman, M. Abdel-Aty; data collection: A. Faden; analysis and interpretation of results: S. Yang, A. Abdelrahman; draft manuscript preparation: A. Abdelrahman, M. Abdel-Aty, S. Yang, A. Faden. All authors reviewed the results and approved the final version of the manuscript.